\documentclass[runningheads,a4paper]{llncs}

\usepackage{ifxetex}
\ifxetex
  \usepackage{fontspec}
  \usepackage{polyglossia}
  \setmainlanguage{english}
\else
  \usepackage[utf8]{inputenc}
  \usepackage[T1]{fontenc}
  \usepackage[english]{babel}
\fi

\usepackage{amssymb}
\usepackage{graphicx}
\usepackage{url}
\usepackage{float}
\usepackage{amsmath}
\usepackage{graphicx}
\usepackage{wrapfig}
\usepackage{booktabs}
\usepackage{multirow}
\usepackage{lipsum}
\usepackage{csquotes}
\usepackage{fancyhdr}
\usepackage{lastpage}
\usepackage[all]{nowidow}
\usepackage[inline]{enumitem}
\usepackage[usenames,dvipsnames]{xcolor}

\makeatletter
\def\Hline{%
\noalign{\ifnum0=`}\fi\hrule \@height 1pt \futurelet
\reserved@a\@xhline}
\makeatother

\begin{document}


%
%
\title{Hibikino-Musashi@Home\\2025 Team Description Paper}

\author{
    Ryohei Kobayashi
    \and Kosei Isomoto
    \and Kosei Yamao
    \and Soma Fumoto
    \and Koshun Arimura
    \and Naoki Yamaguchi
    \and Akinobu Mizutani
    \and Tomoya Shiba
    \and Kouki Kimizuka
    \and Yuta Ohno
    \and Ryo Terashima
    \and Hiromasa Yamaguchi
    \and Tomoaki Fujino
    \and Ryoga Maruno
    \and Wataru Yoshimura
    \and Kazuhito Mine
    \and Tang Phu Thien Nhan
    \and Yuga Yano
    \and Yuichiro Tanaka
    \and Takeshi Nishida
    \and Takashi Morie
    \and Hakaru Tamukoh
}
\authorrunning{Ryohei Kobayashi et al.}
\institute{
Kyushu Institute of Technology\\
The University of Kitakyushu\\
\email{hma@brain.kyutech.ac.jp} \\
\url{https://www.brain.kyutech.ac.jp/~hma/}
}

\maketitle

\begin{abstract} 

This paper provides an overview of the techniques employed by Hibikino-Musashi@Home, which intends to participate in the domestic standard platform league. The team developed a dataset generator for training a robot vision system and an open-source development environment running on a Human Support Robot simulator. The large-language-model-powered task planner selects appropriate primitive skills to perform the task requested by the user. Moreover, the team has focused on research involving brain-inspired memory models for adaptation to individual home environments. This approach aims to provide intuitive and personalized assistance. Additionally, the team contributed to the reusability of the navigation system developed by Pumas in RoboCup2024. The team aimed to design a home service robot to assist humans in their homes and continuously attend competitions to evaluate and improve the developed system.

\end{abstract}

\section{Introduction} 

Hibikino-Musashi@Home (HMA) is a robot development team organized by students at the Kyushu Institute of Technology and the University of Kitakyushu in Japan. The team was founded in 2010 and has participated in the RoboCup, RoboCup JapanOpen, and World Robot Challenge. HMA focuses on the development of a robot vision system, particularly a dataset generation system for training object recognition models. The team also develops libraries for fundamental tasks such as object recognition, grasping point estimation, and navigation. Recently, task planning has become a central research topic, where HMA uses a large language model (LLM) to plan tasks by dynamically selecting primitive actions based on the environment.

\section{Perception}

\subsection{Object Recognition}
\label{ss:object-detection}
Object recognition systems are crucial components of robot systems. We adopted a strategy that leveraged both YOLOv8 \cite{Jocher_YOLO_by_Ultralytics_2023} and a combination of Grounding DINO \cite{G_DINO} and NanoSAM \cite{nano_sam}. We selected the best system for each task depending on the target object.

\subsubsection{YOLO} 
We fine-tuned YOLOv8 to recognize the objects used in the computation by using an enormous training dataset generated by a 3D simulation system based on the PyBullet \cite{coumans2016pybullet} simulator \cite{ono2022ar}. To improve the performance of the model on edge devices, we utilized TensorRT \cite{tensorrt} quantization techniques to reduce the computational complexity. Quantization enables the model to run efficiently on devices with limited computational power, such as real-time robotic systems. To generate the training dataset, we first created a 3D model of each object using a smartphone equipped with LiDAR sensors. The scanned 3D objects were spawned in a 3D environment, and the objects and environment were captured from various angles to create a dataset with 500,000 images in less than two hours using a six-core CPU simultaneously. The lighting conditions, furniture placement, and background texture (floor, wall, and ceiling) were changed to randomize the domain at each shot. The annotation data for the training data can be generated automatically by the system, and no human annotation process is required.
  
\subsubsection{Combination of GroundingDino and NanoSAM} 
The combination of GroundingDino and NanoSAM is a highly efficient recognition system that uses a text prompt and a lightweight segmentation model. Grounding DINO allows for quick adjustments without extensive preparation, rendering it particularly useful for recognizing known objects available on a competition site. The target object can be detected by providing different types of prompts such as colors, materials, and categories. Humans were required to choose and tune the prompt by checking the recognition results in advance.
NanoSAM is a lightweight model that produces an object mask. The image and  bounding box generated by Grunding DINO are input to the model, and the mask used for grasping point estimation is the output.

\begin{figure}[tb]
\begin{center}
    \includegraphics[width=0.5\columnwidth]{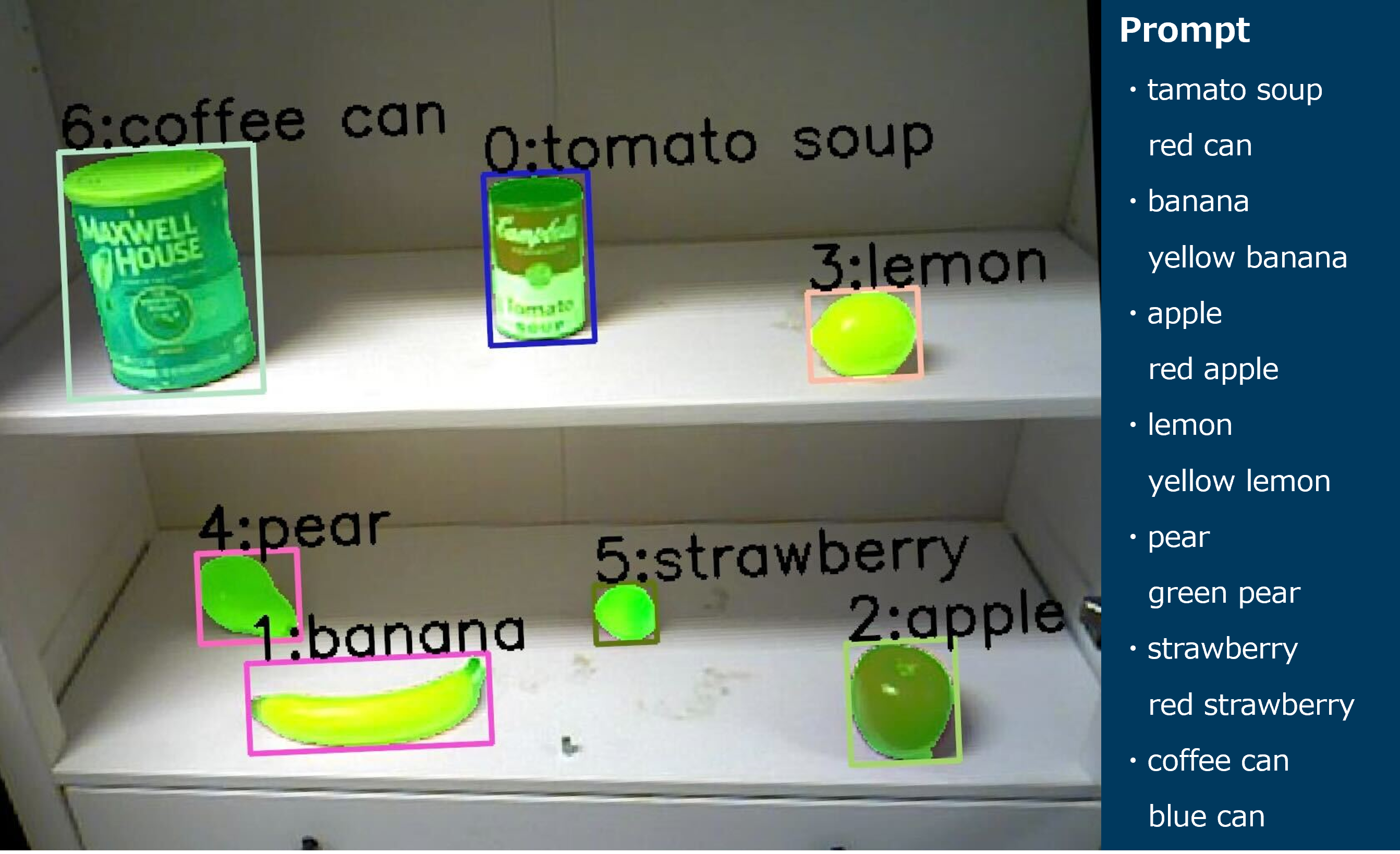}
    \caption{Example of the prompts for GroundingDino + NanoSAM.}
    \label{fig:nanosam_result}
  \end{center}
  \end{figure}

\subsection{Grasping Pose Estimation}
\label{ss:grasping-point} 
\begin{figure}[tb] 
  \begin{center}
    \includegraphics[width=1.0\columnwidth]{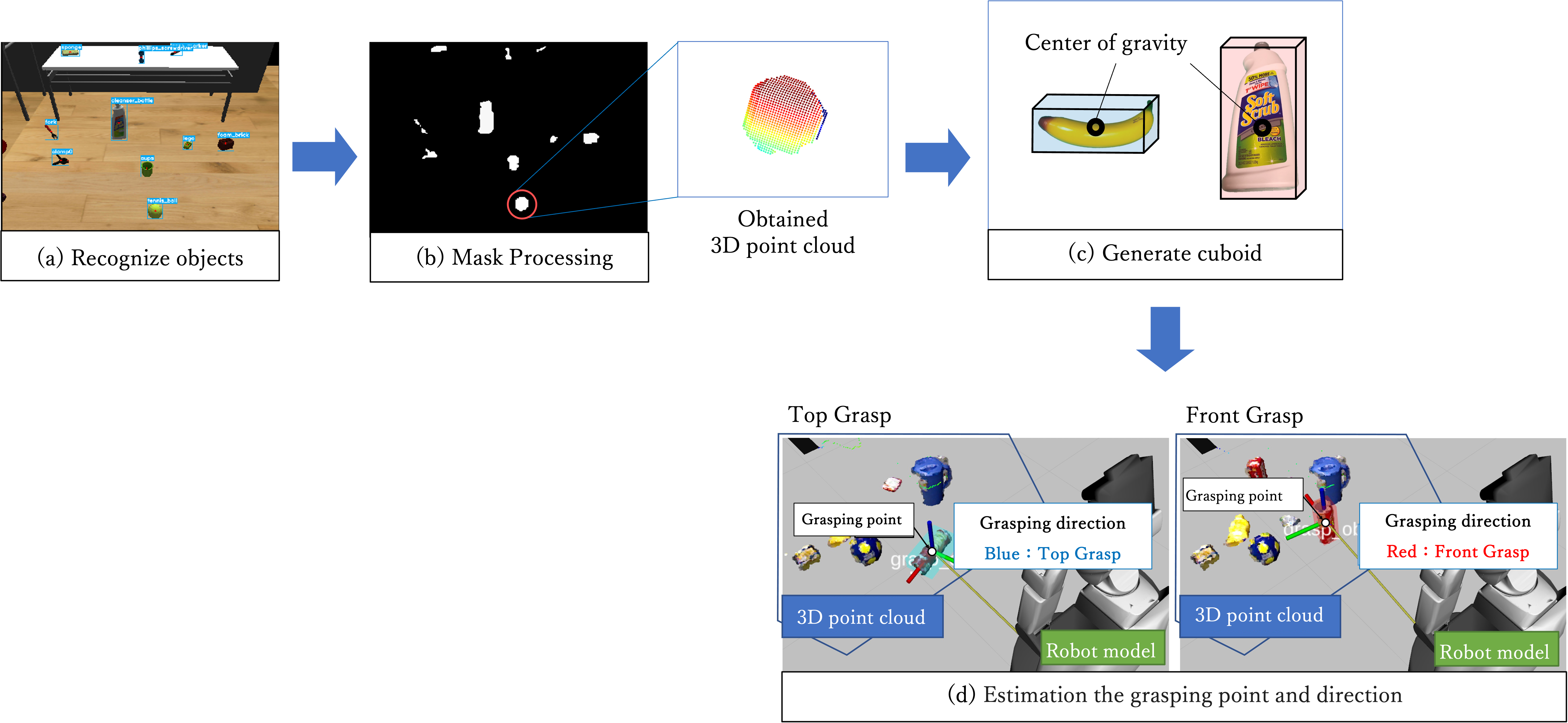}
    \caption{Overview of the grasping point estimation.}
    \label{fig:grasp_point}
  \end{center}
\end{figure}

The robotic system calculates the grasping pose as shown in Fig. \ref{fig:grasp_point} by using a 3D point cloud. The procedure for the proposed method is as follows:

\begin{itemize}
  \setlength{\leftskip}{1.0cm}
  \item [Step (a)] Detect and segment objects using object recognition (YOLO, Ground-Dino + NanoSAM), as described in Subsection \ref{ss:object-detection}.
  \item [Step (b)] Masks the 3D point cloud from the generated mask image and obtains the 3D point cloud of the object closest to the robot.
  \item [Step (c)] Generates a 3D bounding box of the object from the 3D point-cloud data using PCA and calculates its center of gravity.
  \item [Step (d)] Estimates the grasping pose based on the width and height of the 3D bounding box and its center of gravity (e.g., the blue 3D bounding box is grasped from the top because of its long width).
\end{itemize}

Using this method, the robot can flexibly and swiftly estimate the grasping pose using the visual information of the object, regardless of the angle at which the object is placed.

\subsection{Human Action Recognition} 
\begin{figure}[tb]
  \begin{center}
  \includegraphics[width=0.9\columnwidth]{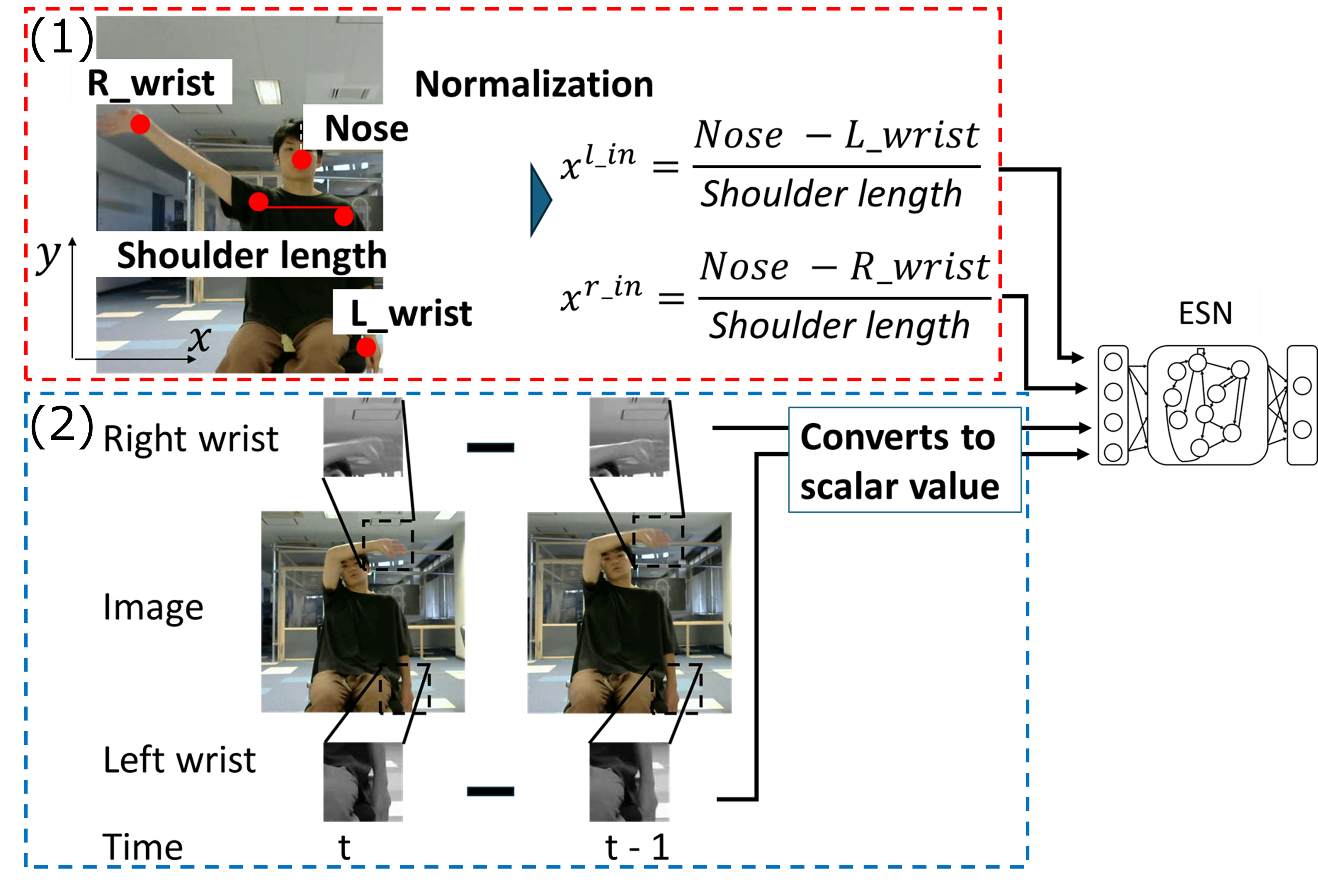} 
  \caption{Overview of the hand waving recognition \cite{hand_waving_recognition}}
  \label{fig:Human_Action_Recognition}
  \end{center}
\end{figure}

We proposed a hand-waving detection system with low power consumption and low latency. It employs a low-computational echo state network (ESN) \cite{jaeger2001echostate} using normalized human skeletal data and fingertip image data as inputs, as illustrated in Fig. \ref{fig:Human_Action_Recognition} \cite{hand_waving_recognition}. Skeletal coordinates were captured using MediaPipe \cite{lugaresi2019mediapipe}. In step (1), Skeletal data were normalized to achieve position-invariant recognition. In step (2), the fingertip area is cropped from the image, and the sum of the pixel values is used as an input, which helps handle cases in which only the fingertip moves while the wrist remains stationary. The system performs binary classification to determine whether a person is waving by inputting normalized skeletal data.
(1) and fingertip image data (2) into the ESN. Compared to conventional deep learning-based action recognition systems, this method achieves equivalent accuracy while reducing the CPU processing time by approximately 15 s.

\subsection{Semantic Map} 
\begin{figure}[tb]
  \begin{center}
    \includegraphics[width= 0.5\columnwidth]{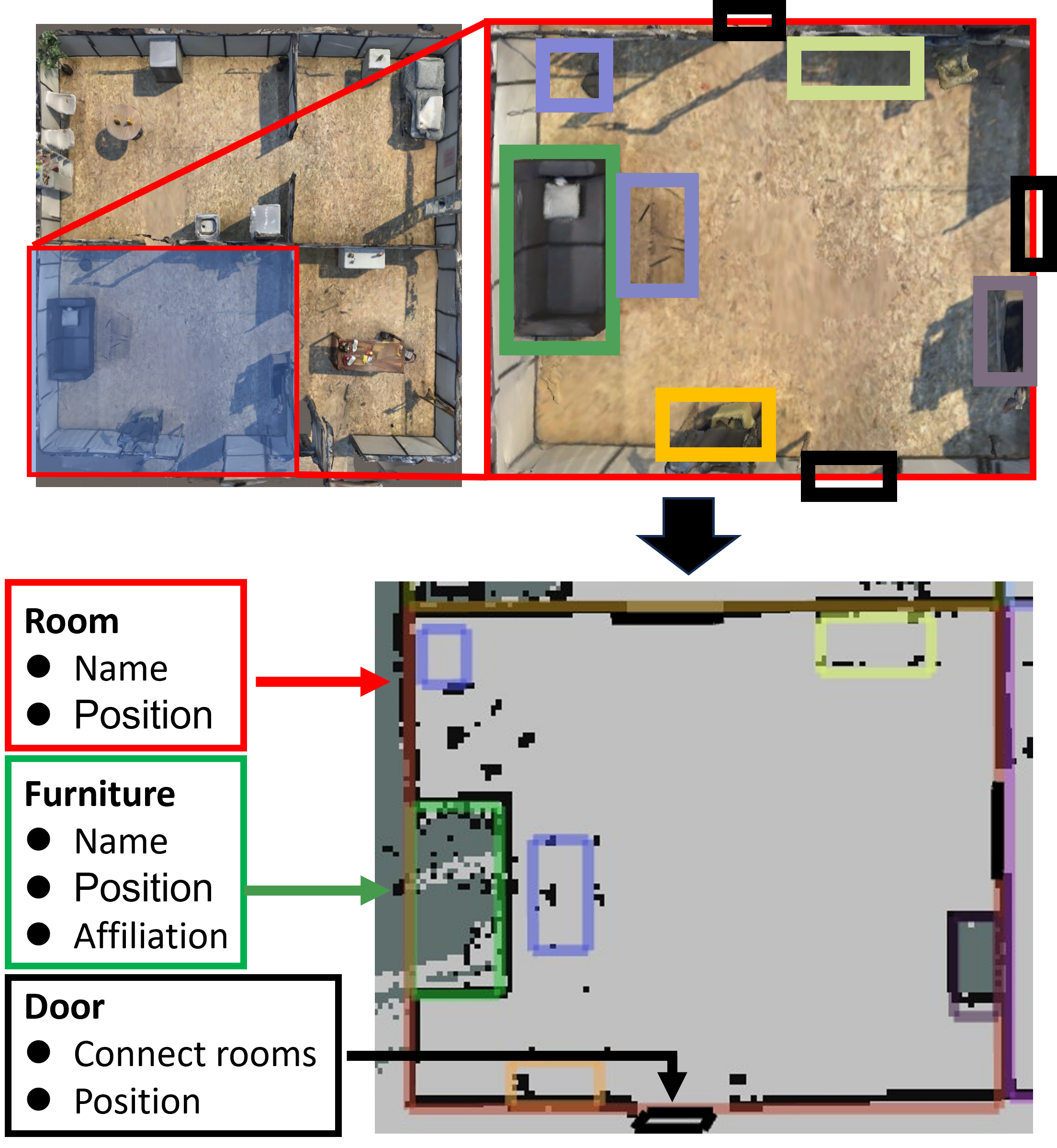}
    \caption{Example of a semantic map}
    \label{fig:semantic-map}
  \end{center}
\end{figure}

Semantic maps link the names of rooms and furniture to location information. As shown in Fig. \ref{fig:semantic-map}, our semantic map adds semantic information about rooms, furniture, and doors to the environmental map. Each room and furniture had a label with a name and location information using an array of two-dimensional (2D) coordinates representing the vertex information of the contours. Moreover, furniture contains information about the room in which it is located, and each door contains information about the room to which it is connected. When visualizing rooms and furniture on the semantic map, a random color was assigned to each area for easier distinction.

The robot uses semantic information to determine the location of the robot or human. The inside/outside determination using the outer product determines the room in which the robot is located based on the contour information of all rooms and the target location. To determine the navigation point, the robot first obtains the name of the target location from the linguistic instructions. A normal is then drawn from the line segment that represents the contour of the target location, and the navigation point is the location at a distance that facilitates recognition and manipulation. Navigation performance can be improved using information on the shape of furniture contained in the semantic map as an obstacle to the environmental map.


\section{Toward General Purpose Service Robots} 
\begin{figure}[tb]
  \begin{center}
  \includegraphics[width=0.9\columnwidth]{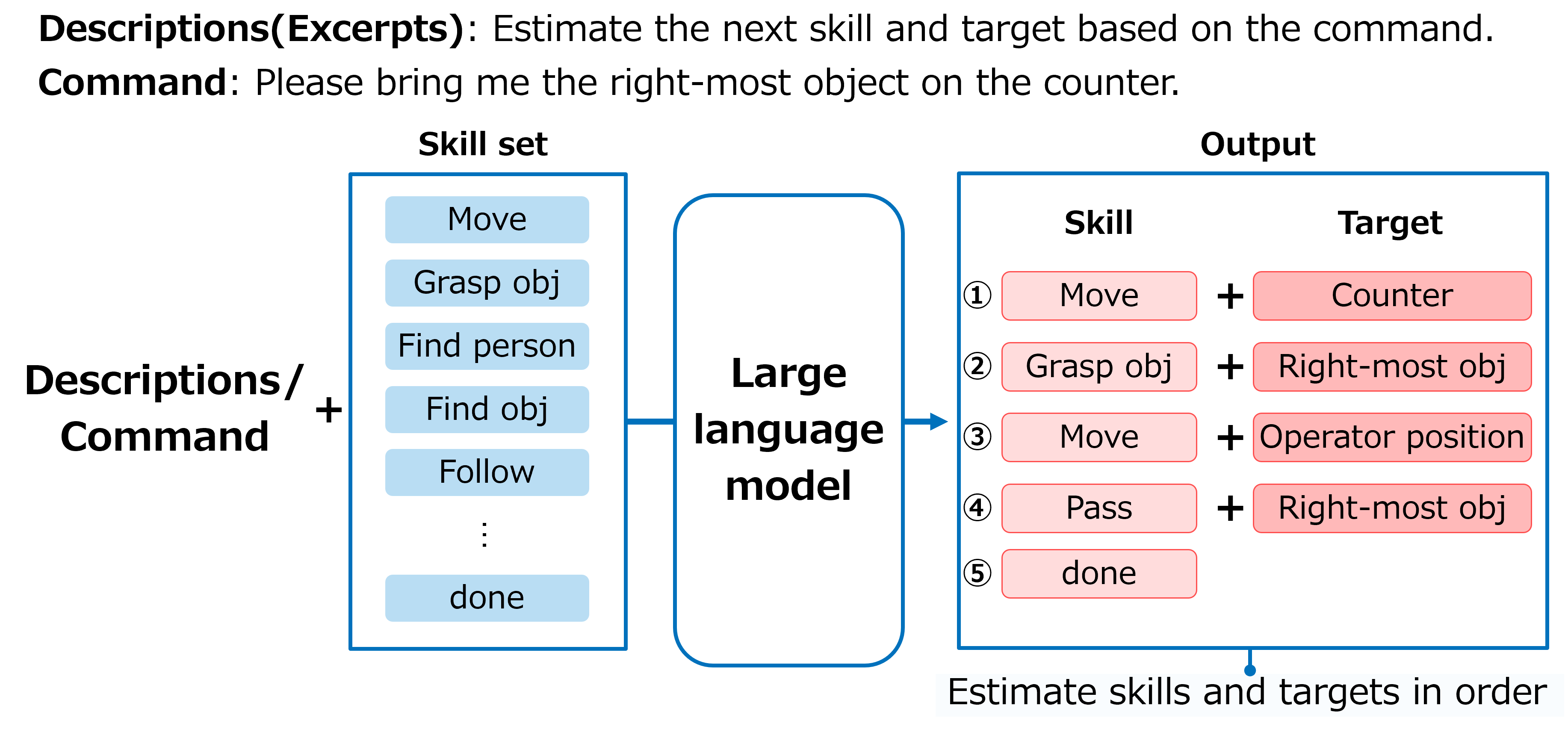} 
  \caption{Overview of the task planning \ref{fig:task_planning}}
  \label{fig:task_planning}
  \end{center}
\end{figure}

A task planning system is required to accomplish various user commands in a real-world environment. As shown in Fig. \ref{fig:task_planning}, a proposed system plans the task using a large language model (LLM) \cite{gpt-4o}. First, the voice commands from the user are converted into text using Whisper \cite{radford2022robust}], a speech recognition model. Subsequently, a description of the robot’s role and other information, commands, and skill sets were entered into the LLM. The skill set is the set of skills available to the robot. The LLM selects appropriate skills from the skill set and determines their targets. The skill set includes 12 skills: \texttt{find\_obj}, \texttt{grasp}, and \texttt{move}. For example, \texttt{find\_obj} is a skill in which the robot detects the target objects using YOLOv8 and Language Segment Anything, as discussed in Section \ref{ss:object-detection}. \texttt{grasp} is a skill in which the robot estimates the grasping point and orientation of a target object based on a 3D point cloud and grasps it as described in Section \ref{ss:grasping-point}. In the case of a command such as \textit{Bring me the right-most object on the counter}, the system sequentially determines the skills and targets until it is completed, as shown in Fig. \ref{fig:task_planning}.

To extend beyond the general-purpose service robot task functions, the robot must autonomously acquire information about its home environment. We focused on the memory function of the brain to form episodic memory and developed an artificial intelligence model inspired by the function of the hippocampus and related areas of the brain \cite{mizutani2024}. This brain-inspired memory model is utilized to memorize episodes in the home, such as a mother buying milk and putting it in a refrigerator the next day or a robot placing a glass of water on the table every day. For example, when a user requests the delivery of an object, the model minimizes the location of the object and utilizes that memory for a future event. A large language model was used to facilitate the acquisition of initial knowledge. The model was tuned to output the potential object locations. We also provided continuously acquired object locations as contextual information. This contextual information contributes to the inference that the novel object location is more suitable for the home environment. A brain-inspired memory model has been implemented in a field-programmable gate array and a dedicated chip for energy-efficient computing \cite{kawashima2022,shishido2024-ijcnn,shishido2024-nolta}. Our future vision for a service robot with an LLM-based task planner and a brain-inspired memory model will be demonstrated in the open challenge and final stages.

\section{Reusability} 
We published our development workspace used in RoboCup 2021 on open-source HSR simulators \cite{hsrb_robocup_dspl_docker} on GitHub\footnote{\texttt{hma\_wrs\_sim\_ws} (\url{https://github.com/Hibikino-Musashi-Home/hma_wrs_sim_ws})}. This includes documentation and sample programs with motion synthesis and object recognition libraries. This simulator workspace enables the development of a robot system even without physical HSR. It can also be used to test and evaluat e robotic systems. Currently, we are developing an open-source workspace for physical HSR with a virtual singularity environment. As a part of this effort, we are releasing and maintaining navigation systems. This navigation system on Github was released by Pumas, a team participating in the RoboCup@Home OPL, and included a Path Planner, Obstacle Detector, and Human Follower \cite{robot_navigation}. We utilized this system for all tasks in the 2024 DSPL and contributed to the maintenance of the system by releasing a version with bug fixes, error handling, and other improvements identified during the development process.

\section{Conclusions} 
This paper outlines techniques for developing an intelligent home service robot system. An automatic dataset-generation system is crucial for training a robot’s visual system within a limited timeframe. The task planning system is sufficiently robust to generate actions based on human voice commands, and the team continually develops essential primitive skills to further enhance the robot’s capabilities.

\section*{Acknowledgement} 
This paper is based on results obtained from a project, JPNP16007 and JPNP20004 commissioned by the New Energy and Industrial Technology Development Organization (NEDO).
This paper is also supported by Joint Graduate School Intelligent Car, Robotics \& AI, Kyushu Institute of Technology student project, YASKAWA electric corporation project, JSPS KAKENHI grant number 23H03468, 23K18495, JST ALCA-Next Grant Number JPMJAN23F3, and JST SPRING grant number JPMJSP2154.


\bibliography{ref}
\bibliographystyle{splncs04} 

\newpage


\section*{Appendix 1: Robot's Software Description} 
\begin{wrapfigure}[20]{r}{0.25\textwidth}
        \centering
        \includegraphics[width=0.25\textwidth]{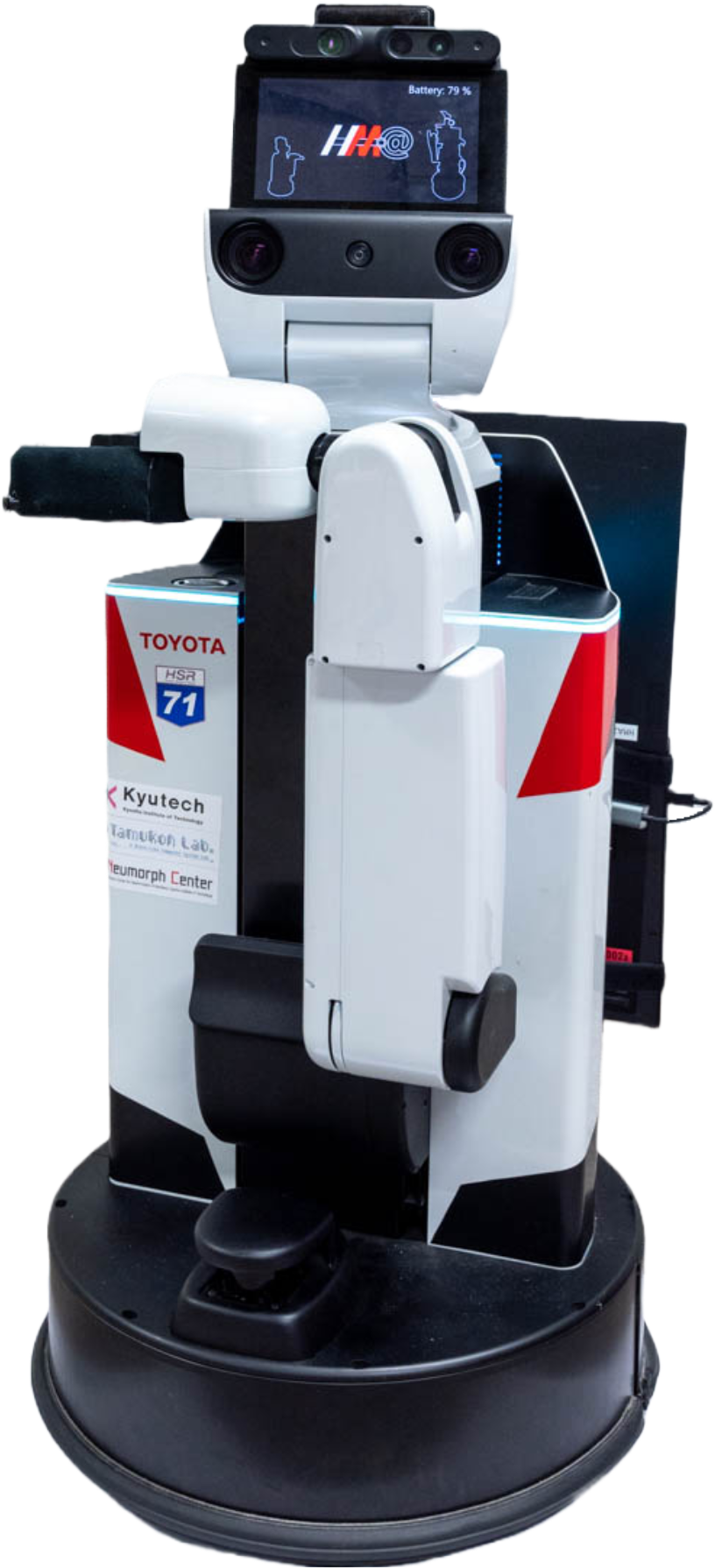}
        \caption{HSR}%
        \label{fig:hsr}
\end{wrapfigure}

The software stack for the robot system is shown in Fig. \ref{fig:hsr}.

\begin{itemize}
	\item Host OS: Ubuntu 22.04
    \item Container: Singularity \cite{singularity}
	\item Middleware: ROS Noetic
	\item State management: SMACH (ROS)
    \item Object detection: YOLO and NanoSam \cite{Jocher_YOLO_by_Ultralytics_2023} \cite{nano_sam}
	\item Speech recognition: Whisper \cite{radford2022robust} and Voice Activity Detection \cite{pyvad2016}
    \item Human detection: OpenPose \cite{Cao2018OpenPose} 
	\item Attribute recognition: CSRA \cite{Zhu2021CSRA}
    \item Human action recognition: A ESN based Human action recognition system \cite{jaeger2001echostate} \cite{hand_waving_recognition}
    \item Face detection: FaceNet \cite{Schroff2015FaceNet}
	\item Mapping: Cartographer \cite{cartographer}
    \item Localization: Adaptive Monte Carlo Localization (AMCL) \cite{probabilistic}
    \item Navigation: Pumas Navigation \cite{negrete:2018} 
\end{itemize}

The following are the specifications of the laptop mounted on our HSR.
\begin{itemize}
        \item Model name: ROG Zephyrus M16 GU604VY
        \item CPU: Intel Core i9-13900H
        \item RAM: 32GB 
        \item GPU: NVIDIA GeForce RTX 4090 (16GB) 
\end{itemize}

\section*{Appendix 2: Competition results} 
Table \ref{tab:result} shows the results achieved by our team during recent competitions. We have been participating in the RoboCup and World Robot Challenges for several years. Our team has won several prizes and academic awards since then.

\begin{table}[t]
\begin{center}
\caption{Results of the recent competitions. [DSPL, domestic standard-platform league; JSAI, Japanese Society for Artificial Intelligence; OPL, open-platform league]}
\label{tab:result}
\begin{tabular}{l|l} \hline
	\multicolumn{1}{c|}{Competition} & \multicolumn{1}{c}{Result} \\ \hline \hline
  
        2017 Nagoya & {\bf @Home DSPL 1st} \\
                                   & @Home OPL 5th \\ \hline

        JapanOpen 2018 Ogaki & @Home DSPL 2nd \\
                                   & \textbf{@Home OPL 1st} \\
                                   & JSAI Award \\ \hline
    
        2018 Montreal & \textbf{@Home DSPL 1st} \\
                                   & P\&G Dishwasher Challenge Award \\ \hline


        2019 Sydney & @Home DSPL 3rd \\ \hline

        JapanOpen 2019 Nagaoka &\textbf{@Home DSPL 1st} \\
                                        & \textbf{@Home OPL 1st} \\ \hline
        JapanOpen 2020 & @Home Simulation Technical Challenge 2nd \\
                                        & \textbf{@Home DSPL 1st} \\
                                        & @Home DSPL Technical Challenge 2nd \\
                                        & \textbf{@Home OPL 1st} \\
                                        & \textbf{@Home OPL Technical Challenge 1st} \\
                                        & @Home Simulation DSPL 2nd \\ \hline
        Worldwide 2021 & @Home DSPL 2nd \\
                                        & \textbf{@Home Best Open Challenge Award 1st} \\
                                        & \textbf{@Home Best Test Performance: } \\
                                        & \textbf{    Go, Get It! 1st} \\
                                        & \textbf{@Home Best Go, Get It! 1st} \\ \hline
        Asia-Pacific 2021 Aichi & \textbf{@Home DSPL 1st}\\
                                        & \textbf{@Home OPL 1st} \\ \hline
        JapanOpen 2021 & \textbf{@Home DSPL 1st} \\
                                        & \textbf{@Home DSPL Technical Challenge 1st} \\
                                        & @Home OPL 2nd \\
                                        & \textbf{@Home OPL Technical Challenge 1st} \\ \hline
        2022 Bangkok & @Home DSPL 3rd\\ 
                                        & @Home Best Open Challenge Award\\
                                        & @Home Robo-host \\
                                        & (Party-Host highest score in Stage I tasks)\\ \hline
        JapanOpen 2022 Tokyo & \textbf{@Home DSPL 1st} \\
                                        & @Home DSPL Technical Challenge 2nd \\
                                        & @Home OPL 2nd \\
                                        & \textbf{@Home OPL Technical Challenge 1st} \\ \hline
        JapanOpen 2023 Shiga & \textbf@Home DSPL 3rd \\
                                        & \textbf{@Home DSPL Open Challenge 1st} \\
                                        & @Home OPL 2nd \\
                                        & \textbf{@Home OPL Open Challenge 1st} \\ \hline
        2023 Bordeaux & @Home DSPL 2nd\\ \hline
	JapanOpen 2024 Shiga & @Home DSPL 2nd \\
					& @Home DSPL Open Challenge 2nd \\
					& @Home OPL 2nd \\
					& \textbf{@Home OPL Open Challenge 1st} \\
					& @Home Simulation OPL 2nd \\ \hline
	2024 Eindhoven & \textbf{@Home DSPL 1st} \\
                                        & @Home GPSR Overbot\\
                                        & (Best in GPSR \& EGPSR) 1st\\
                                        & @Home Smoothest, Safest Navigation Award 1st\\
                                        & @Home Robo-host\\
                                        & (Party-Host highest score in Stage I tasks) 1st\\
                                        & @Home Robo-butler\\
                                        & (Housekeeper highest score in Stage II tasks) 1st\\ \hline
\end{tabular}
\end{center}
\end{table}

\section*{Appendix 3: Links}

\begin{itemize}
  \item Team Video \\ \url{https://youtu.be/73GkRklouho} \\
  \item Team Website \\ \url{https://www.brain.kyutech.ac.jp/~hma} \\
  \item GitHub \\ \url{https://github.com/Hibikino-Musashi-Home} \\
  \item Facebook \\ \url{https://www.facebook.com/HibikinoMusashiAthome} \\
  \item YouTube \\ \url{https://www.youtube.com/@hma_wakamatsu}
\end{itemize}

\end{document}